\documentclass{amsart}
\usepackage{amssymb}
\usepackage{geometry}
\usepackage{hyperref}
\usepackage{algorithm}
\usepackage{algpseudocode}
\usepackage{tikz}
\usepackage{pgfplots}
\pgfplotsset{compat=1.18}

\usepackage[numbers]{natbib}
\setcitestyle{numbers,sort&compress}
\AtBeginEnvironment{thebibliography}{\small}

\begin{document}

\title{New Hybrid Heuristics for Pseudo-Boolean Propagation}

\author{Mia Müßig}
\address[Mia Müßig]{Institut für Informatik, Ludwig-Maximilians-Universität München, Germany}
\email{nienna@miamuessig.de}

\author{Jan Johannsen}
\address[Jan Johannsen]{Institut für Informatik, Ludwig-Maximilians-Universität München, Germany}
\email{jan.johannsen@ifi.lmu.de}

\begin{abstract}
In pseudo-boolean solving the currently most successful unit propagation strategy is a hybrid mode combining the watched literal scheme with the counting method. This short paper introduces new heuristics for this hybrid decision, which are able to drastically outperform the current method in the \textsc{RoundingSAT} solver.
\end{abstract}

\maketitle

\section{Introduction}

Pseudo-boolean solvers are an extension of SAT solvers that allow general linear inequalities to be used as constraints and optionally a linear objective function to be specified. This increased flexibility for users comes at the cost of complicating many components of the CDCL algorithm, with both conflict analysis and unit propagation still being continuously improved by ongoing research \cite{elffers,lomis,nieu25}.

Here, we focus on the latter, where the simple counting method was the dominant paradigm until 2020, when Devriendt \cite{devriendt} developed a specialised watched literal scheme for the \textsc{RoundingSAT} solver. This implementation was further improved by Nieuwenhuis et al. \cite{nieuwenhuis}, who demonstrated that the existing hybrid method combining the counting and watched literal scheme could outperform both approaches. Finally, the authors of this paper recently extended the watched literal scheme with the significant literal approach \cite{muessig}, which could improve the runtime for specific applications like knapsack instances.

In this paper we will apply the most successful heuristics from the significant literal scheme to the hybrid method to improve the constraint-specific choice between watched literals and the counting approach. We observe a remarkable improvement across the decision and optimization instances from recent competition datasets.

\section{Preliminaries}

A \textit{pseudo-boolean problem} consists of variables $x_{i} \in \{0, 1\}$, with \textit{literals} $l_{i}$ representing either $x_{i}$ or $\overline{x}_{i} := 1 - x_{i}$ and \textit{pseudo-boolean constraints} $C$ as $\sum_{i = 1}^n a_{i} l_{i} \geq b$ for $a_{i}, b \in \mathbb{N}$. Without loss of generality we will assume that the coefficients are in descending order, so $\forall i : a_{i} \geq a_{i + 1}$. The \textit{slack} of a pseudo-boolean constraint under the current (partial) assignment $\rho$ of the literals is defined as $slack(C, \rho) = -b + \sum_{\overline{l}_{i} \notin \rho} a_{i}$, which is the amount by which the left-hand side can exceed the right-hand side. Thus, if $slack(C, \rho) < 0$, the constraint $C$ is unsatisfiable without unassigning literals. A literal $l_{i}$ is called a \textit{unit literal} if $slack(C, \rho) < a_{i}$. This means that the literal $l_{i}$ must be set to $1$ as otherwise $slack(C, \rho \cup \{\overline{l_{i}}\}) < 0$.

In the \textit{counting method} for detecting unit literals, each time we assign or unassign a variable, we update the slack of all constraints in which it occurs. The \textit{watched literal scheme} by Devriendt instead maintains a set of literals $W(C)$ for each constraint, which all must be either set to $1$ or unassigned and satisfy the following inequality:
\begin{equation}
     \sum_{l_{i} \in W(C)} a_{i} \geq b + a_{1}\label{eq:watch}
\end{equation}

Similar to the watched scheme for SAT we now only have updates to $W(C)$ when assigning variables and only in constraints where the variable is a watched literal. Contrary to SAT, where using watched literals has been the dominant method for more than two decades, in pseudo-boolean solving this method only offers a slight improvement over the counting approach. We refer the reader to Devriendts paper \cite{devriendt} for more details on the mathematical background and important implementation optimizations.

The current implementation of the \textsc{RoundingSAT} solver uses a hybrid method between the counting and watched literal scheme \cite{roundingsat}. Each time a new constraint is added, either as part of the original instance or via conflict analysis, algorithm \ref{alg:hybrid} is used to decide if the constraint will be treated using the counting method or otherwise with the watched literal scheme. The parameter $p$ is set with the \textit{prop-counting} option and has a default value of $0.7$.

\begin{algorithm}
  \caption{\textsc{Hybrid} (Constraint \( \sum_{i = 1}^n a_{i} l_{i} \geq b \), propagation parameter \( p \))}
  \label{alg:hybrid}
  \begin{algorithmic}[1]
  \State $sum \gets -b$
  \State $m \gets 1$
  \While{$m < n$ \textbf{and} $sum < 0$}
      \State $sum \gets sum + a_{m + 1}$
      \State $m \gets m + 1$
  \EndWhile
  \State $useCounting \gets (p = 1) \lor \left(p > \left(1 - \frac{m}{n}\right)\right)$
  \end{algorithmic}
\end{algorithm}

The idea is to use $m$ to count the minimal number of watched literal necessary. Since the literals are sorted by descending coefficient, we simply choose the first literals until we fulfill equation \ref{eq:watch}. Because the code initializes $sum \gets -b$ and skips $a_{1}$, the equation is equivalent to $sum \geq 0$. So for the default value 0.7 the code checks if the watched set $W(C)$ must contain $\geq 30\%$ of the literals in $C$, in which case it uses the counting method instead.

\section{New Hybrid Heuristics}

Inspired by the successful heuristics from the significant literal approach \cite{muessig}, we introduce the \textit{Absolute Hybrid Heuristic}:
\begin{equation*}
  useCounting \gets (a_{1} > c)
\end{equation*}
and the \textit{Additive Hybrid Heuristic}:
\begin{equation*}
  useCounting \gets (a_{1} > c + a_{2})
\end{equation*}

We implemented these heuristics with just a few lines of code in commit a5c28f99 of the \textsc{RoundingSAT} solver, which is the newest version at the time of writing. For our benchmark, we use the 1157 instances from the OPT-LIN and DEC-LIN track of the Pseudo Boolean Competition 2025 \cite{PBCompetition2025}. However, we filter out instances where all coefficients of all input constraints are smaller than $100$, which we will call \textit{small} instances. The reason is that both our heuristics only perform well for $c \geq 100$, so for the small instances $useCounting$ would simply be always false and thus the performance identical to the current \textsc{RoundingSAT} solver without hybrid mode. Additionally, this leaves us with a more manageable dataset of 206 instances, which we run with a $3600s$ timeout on an ``AMD Ryzen 7 7735HS'' CPU and 16 GB of memory. We run the current hybrid mode with its default value of $0.7$, but also with $0.8$ to check if a larger cut-off value performs better on the large instances. For all measurements, the optional SoPlex LP solver is enabled. Our implementation, the benchmark scripts and all measurements are available on Gitlab \cite{gitlab}.

\begin{figure}[h!]
    \centering
    {\footnotesize
    \begin{tikzpicture}
      \begin{axis}[
          xmin=70,
          xmax=100,
          ymin=0,
          ymax=3600,
          ytick={0, 500, 1000, 1500, 2000, 2500, 3000, 3500},
          yticklabels={0s, 500s, 1000s, 1500s, 2000s, 2500s, 3000s, 3500s},
          grid=major,
          legend pos=north west,
          legend style={font=\tiny, inner xsep=2pt, inner ysep=2pt, scale=0.8},
          width=11cm,
          height=9cm
      ]

      \addplot[black, thick, mark=*, mark size=2pt] table {data/runtime-original.dat};
      \addplot[purple, thick, densely dashdotted, mark=hexagon*, mark size=3pt] table {data/runtime-counting.dat};
      \addplot[cyan, thick, loosely dashed, mark=otimes*, mark size=2.5pt] table {data/runtime-hybrid.dat};
      \addplot[brown, thick, densely dashed, mark=star, mark size=2.5pt] table {data/runtime-hybrid-0.8.dat};

      \addplot[green, thick, dashed, mark=square*, mark size=2pt] table {data/runtime-add-500.dat};
      \addplot[blue, thick, mark=triangle*, mark size=2.5pt] table {data/runtime-add-1000.dat};

      \addplot[red, thick, dashdotted, mark=diamond*, mark size=2.5pt] table {data/runtime-abs-500.dat};
      \addplot[orange, thick, dotted, mark=pentagon*, mark size=2.5pt] table {data/runtime-abs-1000.dat};
      
      \legend{Watched Literals (91), Counting Method (92), Default Hybrid (96), Default Hybrid 0.8 (93), $a_{1} \geq 500 + a_{2}$ (96),$a_{1} \geq 1000 + a_{2}$ (97), $a_{1} \geq 500$ (95), $a_{1} \geq 1000$ (97)}
      \end{axis}

    \end{tikzpicture}
    \caption{Runtime comparison on the 206 instances of the OPT-LIN and DEC-LIN track from the Pseudo Boolean Competition 2025, where at least one constraint contains a coefficient $\geq 100$.}
    \label{fig:runtime}
    }
\end{figure}
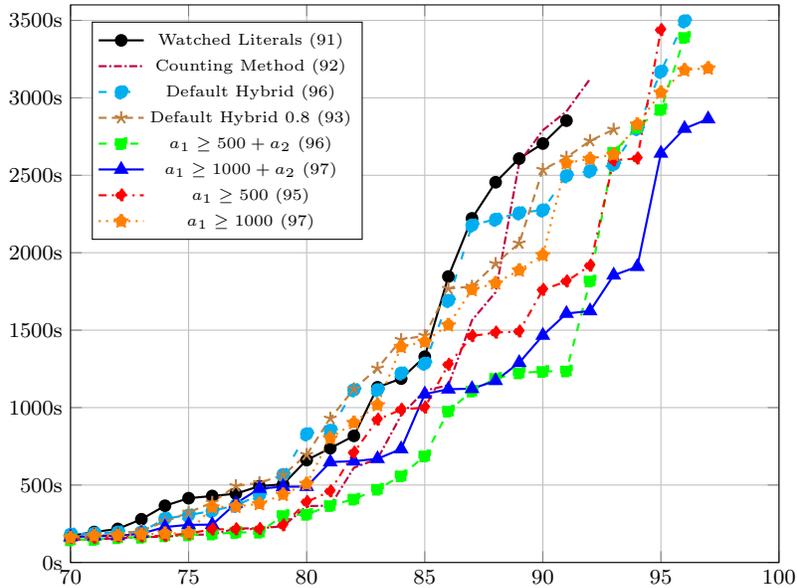

Our first observation from the cactus plot in Figure \ref{fig:runtime} is that the counting method performs slightly better than the watched literal scheme, which is often the case for instances with large coefficients. We can also clearly see that the default hybrid method outperforms the two pure methods it combines. Additionally, choosing a the larger cut-off value of 0.8 for the default hybrid mode leads to a performance degradation. But most importantly, our two new hybrid heuristics perform dramatically better than the current hybrid method. Especially the additive hybrid heuristic with $c = 500$ is vastly faster on average.

\begin{figure}
    \centering
    {\footnotesize
    \begin{tikzpicture}
      \begin{axis}[
          xmin=710,
          xmax=740,
          ymin=0,
          ymax=3600,
          ytick={0, 500, 1000, 1500, 2000, 2500, 3000, 3500},
          yticklabels={0s, 500s, 1000s, 1500s, 2000s, 2500s, 3000s, 3500s},
          grid=major,
          legend pos=north west,
          legend style={font=\tiny, inner xsep=2pt, inner ysep=2pt, scale=0.8},
          width=11cm,
          height=9cm
      ]
      
       Plot all data files with distinct colors
      \addplot[black, thick, mark=*, mark size=2pt] table {data-knap/runtime-original.dat};
      \addplot[purple, thick, densely dashdotted, mark=hexagon*, mark size=3pt] table {data-knap/runtime-counting.dat};
      \addplot[cyan, thick, loosely dashed, mark=otimes*, mark size=2.5pt] table {data-knap/runtime-hybrid.dat};

      \addplot[green, thick, dashed, mark=square*, mark size=2pt] table {data-knap/runtime-add-500.dat};
      \addplot[blue, thick, mark=triangle*, mark size=2.5pt] table {data-knap/runtime-add-1000.dat};
      
      \legend{Watched Literals (737), Counting Method (730), Default Hybrid (735), $a_{1} \geq 500 + a_{2}$ (736), $a_{1} \geq 500 + a_{2}$ (736)}
      \end{axis}

    \end{tikzpicture}
    \caption{Runtime comparison on the 783 knapsack instances from the Pseudo Boolean Competition dataset.}
    \label{fig:runtime-knap}
    }
\end{figure}
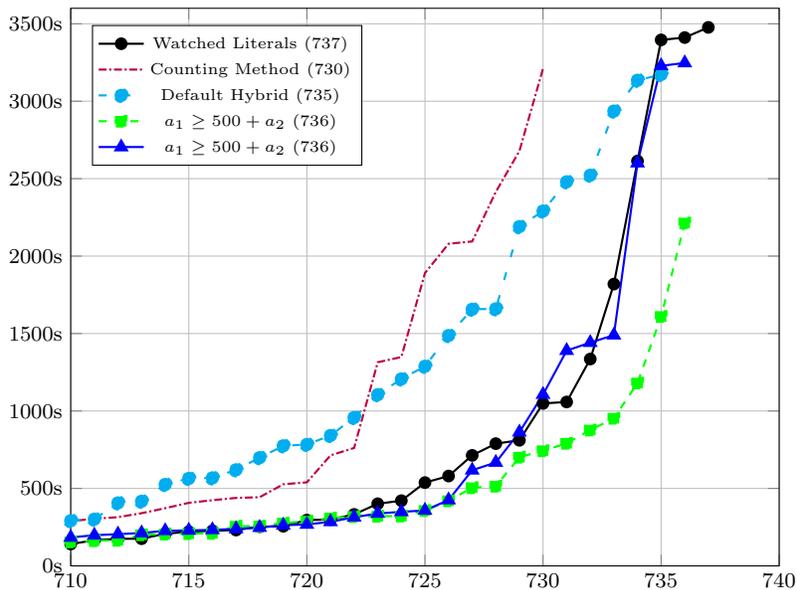

Another dataset that is often used in pseudo-boolean solving research \cite{lp-int, lomis, muessig} are the 783 knapsack instances submitted to the Pseudo Boolean Competition 2024. In Figure \ref{fig:runtime-knap}, it can be observed that criterion $a_{1} \geq 500 + a_{2}$, which performed very well on general instances, again shows a big improvement in average runtime. This result is particularly noteworthy, as in this dataset the current hybrid method performs noticeably worse than the Watched Literal scheme. Again, the data for other values of $c$ and the less effective additive hybrid heuristic can be found on Gitlab.

\section{Conclusion}

We have successfully demonstrated that our proposed change of the hybrid heuristic leads to a significant performance improvement of \textsc{RoundingSAT} and would require only minimal modifications of the source code. Since the classification of which instance is small can be performed in negligible time, one could automatically use the current hybrid heuristic for small instances and the new heuristics for the remaining ones.

\bibliographystyle{ACM-Reference-Format}
\bibliography{paper}

\section{Appendix}

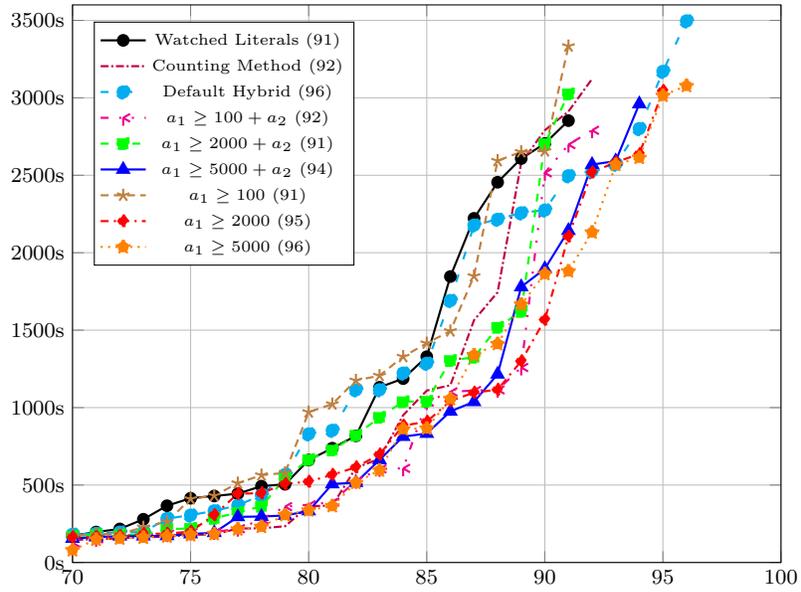
\begin{figure}
    \centering
    {\footnotesize
    \begin{tikzpicture}
      \begin{axis}[
          xmin=70,
          xmax=100,
          ymin=0,
          ymax=3600,
          ytick={0, 500, 1000, 1500, 2000, 2500, 3000, 3500},
          yticklabels={0s, 500s, 1000s, 1500s, 2000s, 2500s, 3000s, 3500s},
          grid=major,
          legend pos=north west,
          legend style={font=\tiny, inner xsep=2pt, inner ysep=2pt, scale=0.8},
          width=11cm,
          height=9cm
      ]

      \addplot[black, thick, mark=*, mark size=2pt] table {data/runtime-original.dat};
      \addplot[purple, thick, densely dashdotted, mark=hexagon*, mark size=3pt] table {data/runtime-counting.dat};
      \addplot[cyan, thick, loosely dashed, mark=otimes*, mark size=2.5pt] table {data/runtime-hybrid.dat};

      \addplot[magenta, thick, loosely dashdotdotted, mark=asterisk, mark size=3pt] table {data/runtime-add-100.dat};
      \addplot[green, thick, dashed, mark=square*, mark size=2pt] table {data/runtime-add-2000.dat};
      \addplot[blue, thick, mark=triangle*, mark size=2.5pt] table {data/runtime-add-5000.dat};

      \addplot[brown, thick, densely dashed, mark=star, mark size=2.5pt] table {data/runtime-abs-100.dat};
      \addplot[red, thick, dashdotted, mark=diamond*, mark size=2.5pt] table {data/runtime-abs-2000.dat};
      \addplot[orange, thick, dotted, mark=pentagon*, mark size=2.5pt] table {data/runtime-abs-5000.dat};
      
      \legend{Watched Literals (91), Counting Method (92), Default Hybrid (96), $a_{1} \geq 100 + a_{2}$ (92), $a_{1} \geq 2000 + a_{2}$ (91),$a_{1} \geq 5000 + a_{2}$ (94),$a_{1} \geq 100$ (91), $a_{1} \geq 2000$ (95), $a_{1} \geq 5000$ (96)}
      \end{axis}

    \end{tikzpicture}
    \caption{Runtime comparison with more $c$ values on the 206 instances of the OPT-LIN and DEC-LIN track from the Pseudo Boolean Competition 2025, where at least one constraint contains a coefficient $\geq 100$.}
    \label{fig:runtime-other}
    }
\end{figure}

In Figure \ref{fig:runtime-other} we show the runtimes for larger and smaller values of $c$. They still outperform the current hybrid mode, but perform worse than $c = 500$ and $c = 1000$. We also experimented with heuristics of the form $a_{1} > c \cdot a_{2}$ and $c > (\max_{1 \leq i < n} a_i - a_{i+1})$, which could not outperform the current hybrid method but are still available on Gitlab \cite{gitlab}.

\end{document}